\pdfoutput=1
\documentclass[11pt,a4paper]{article}

\usepackage[a4paper,left=2cm,right=2cm,top=1cm,bottom=1.25cm,%
            includehead,headheight=50pt,headsep=10pt]{geometry}

\usepackage[T1]{fontenc}
\usepackage[utf8]{inputenc}
\usepackage{cite}
\usepackage{helvet}                        %
\usepackage{xcolor}

\newcommand{\eg}{\emph{e.g.},}

\usepackage{wrapfig}
\usepackage{comment}

\usepackage{setspace}
\usepackage{ragged2e}          %
\usepackage{graphicx}
\usepackage{amsmath,amssymb}
\usepackage{booktabs}
\usepackage{caption}
\usepackage{enumitem}
\usepackage[hidelinks]{hyperref}
\usepackage{caption}
\usepackage{subcaption}
\usepackage{makecell}
\usepackage{bm}

\usepackage{titlesec}
\titleformat{\section}{\normalfont\bfseries}{}{0pt}{}
\titlespacing*{\section}{0pt}{12pt}{8pt}
\titleformat{\subsection}{\normalfont\bfseries}{}{0pt}{}
\titlespacing*{\subsection}{0pt}{8pt}{3pt}
\usepackage{fancyhdr}
\begin{document}

\begin{center}
{\bfseries\large Autonomous 3D Visual-Thermal Inspection for Utility-Scale Solar Farms}
\end{center}

\begin{center}
Lachlan Tychsen-Smith, Mark Cox, Brett Wood, David Haddon, Kenrick Anderson, \\ Benjamin Duck, Jim O'Malley, Noel W Duffy, Peyman Moghadam

CSIRO

\textit{firstname.lastname@csiro.au}
\end{center}

\vspace{-6pt}

\section{Abstract}
\vspace{-2pt}
Utility-scale photovoltaic (PV) inspection is dominated by aerial surveys, which are effective for identifying module-level hot cells but provide limited visibility of critical components located under the array, including connectors, cabling, fasteners and structural elements. These assets are difficult to inspect consistently at scale, and their defects are often spatially small, partially occluded and difficult to localise from isolated images. We present an automated 3D visual-thermal inspection system (\textit{Tricorder}) for under-panel PV asset inspection. The approach combines LiDAR-inertial Simultaneous Localization and Mapping (SLAM) with geometrically calibrated RGB and thermal imagery and machine-learning-based object detection and segmentation. Detected assets are associated with the reconstructed 3D environment, allowing visual and thermal observations from multiple viewpoints to be spatially fused and localised. We demonstrate the approach on utility-scale PV data for MC4 connector detection and connection-state recognition, thermal anomaly identification, and detection of structural components including bolts, nuts and torque tubes. The results demonstrate a pathway from image-based inspection toward spatially grounded autonomous inspection, where defects are detected and linked to specific assets and their 3D locations within the solar farm. %
\section{Problem Statement}
\vspace{-2pt}
Aerial RGB and thermal inspection provides an efficient means of surveying large PV fields and identifying module-level anomalies such as hotspots. However, many critical electrical and structural failure points, including MC4 connectors, cabling, fasteners, and mounting components, are located beneath the modules and are difficult to observe from above. Inspecting these components, therefore, remains dependent on labour-intensive hand-held inspection.
The challenge is not only detecting these assets, but determining what component is being observed, its condition, and where it is located in 3D space. Conventional 2D thermography lacks explicit scene geometry and can be sensitive to viewpoint, occlusion, and thermal reflections. Prior 3D thermography systems demonstrated that combining 3D geometry with visual and thermal observations can improve localisation, repeatability, and multi-view thermal interpretation \cite{moghadam20183d, VIDAS2013445, HeatWave2014}. We extend this principle toward automated, machine-learning-driven inspection of utility-scale PV infrastructure.

\section{Autonomous 3D Visual-Thermal Inspection System}
We formulate under-panel PV inspection as a spatially grounded multimodal perception problem. %
The objective is not only to detect an asset or anomaly in an individual image, but to associate visual and thermal evidence with a specific physical component and its 3D location in the solar farm, enabling candidate faults to be localised.

The proposed system combines four stages. First, LiDAR-inertial SLAM \cite{Wildcat:2022} estimates the sensor trajectory and reconstructs the surrounding three-dimensional environment. Second, geometrically calibrated RGB and thermal observations are aligned to this common spatial reference. Third, machine-learning models \cite{tychsen-smith:2017, tychsen-smith:2018} detect and segment electrical and structural assets within the RGB imagery. Finally, these detections and their associated thermal measurements are projected into the reconstructed 3D map, allowing asset condition and defect observations from multiple viewpoints to be associated with specific physical components.
\section{Tricorder Platform and Data Collection}

\begin{wrapfigure}[14]{r}{0.42\textwidth}
  \vspace{-20pt}
  \centering
  \includegraphics[width=0.30\textwidth]{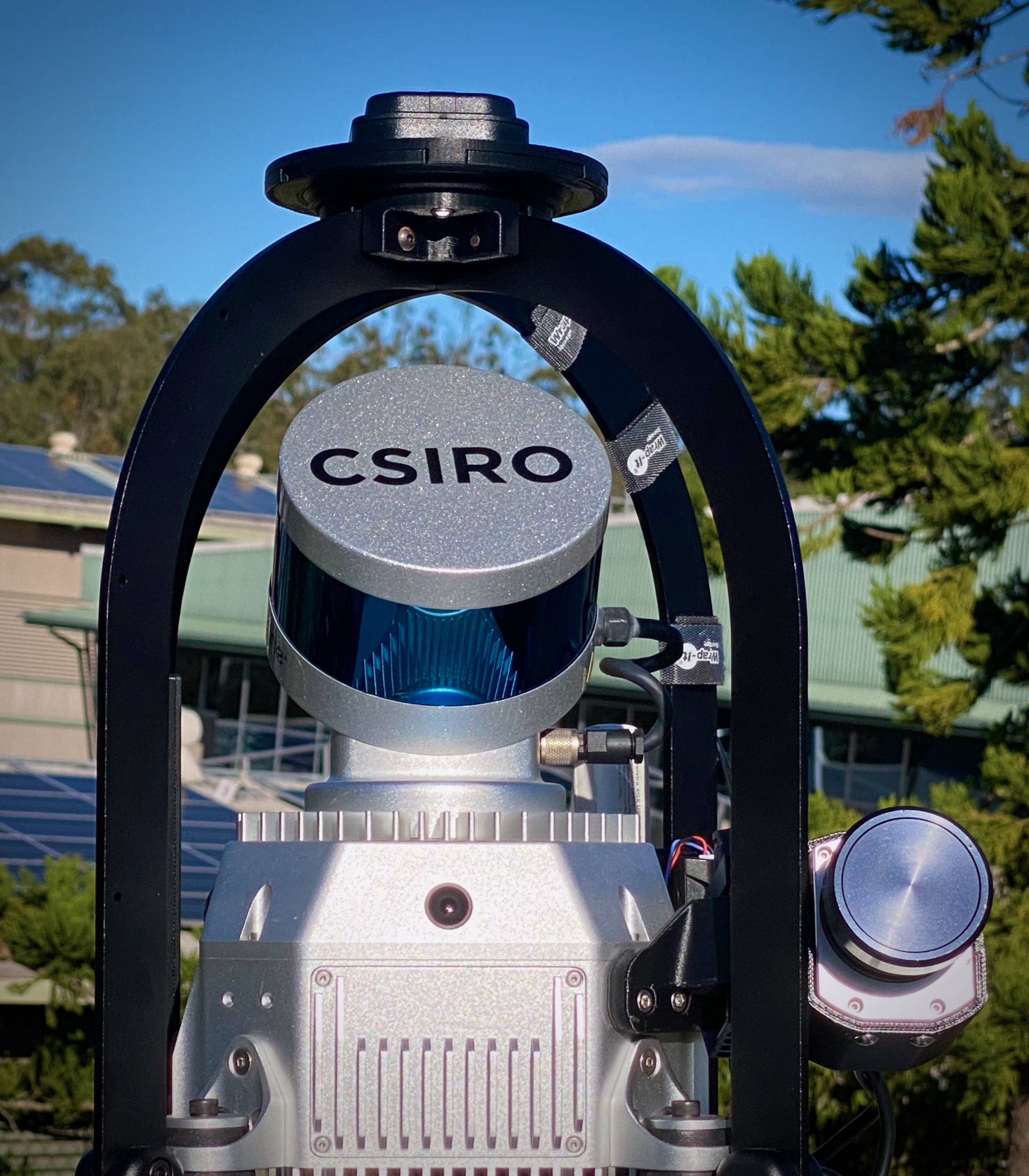}
  \caption{\textit{Tricorder:} a multi-modal sensing platform.}
  \label{fig:device}
\end{wrapfigure}

The sensing platform (called \textit{Tricorder}) integrates a LiDAR sensor, inertial measurement unit (IMU), RGB camera, thermal infrared camera, real-time kinematic (RTK) Global Navigation Satellite System (GNSS) receiver, and an NVIDIA Jetson Xavier for onboard data capture and machine learning processing. The platform can be operated in either handheld or vehicle-mounted configurations, enabling rapid automated inspection of large-scale solar installations. LiDAR, IMU, and GNSS measurements are used to generate a georeferenced three-dimensional representation of the solar farm. The RGB and thermal cameras are geometrically calibrated with respect to the LiDAR coordinate frame, allowing any asset detections and thermal measurements to be accurately localised within a global coordinate system.

\section{Field Deployment and Experimental Evaluation}

Field data were collected at CSIRO's Photovoltaic Outdoor Research Facility in Newcastle, and three commercial utility-scale PV solar farms around Australia. These deployments capture substantial variation in PV structure, component configuration, illumination, weathering, corrosion and background clutter, providing representative data for developing and evaluating under-panel asset detection and defect-identification methods.

A subset of the collected RGB imagery was manually annotated with electrical and structural asset classes, including MC4 connectors and their connection state, bolts, nuts, bolt holes, torque marks, torque tubes and cabling. These annotations were used to train the DeNeT deep-learning framework for object detection and semantic segmentation. %
We demonstrate this capability through two representative use cases: structural inspection for missing fasteners and electrical inspection of MC4 connectors. %

\subsection{Structural Inspection: Missing Fasteners}

\begin{figure}[h]
  \centering
  \begin{subfigure}[t]{0.32\textwidth}
    \centering
    \includegraphics[trim={0 4.5cm 13.2cm 0},clip,height=4cm]{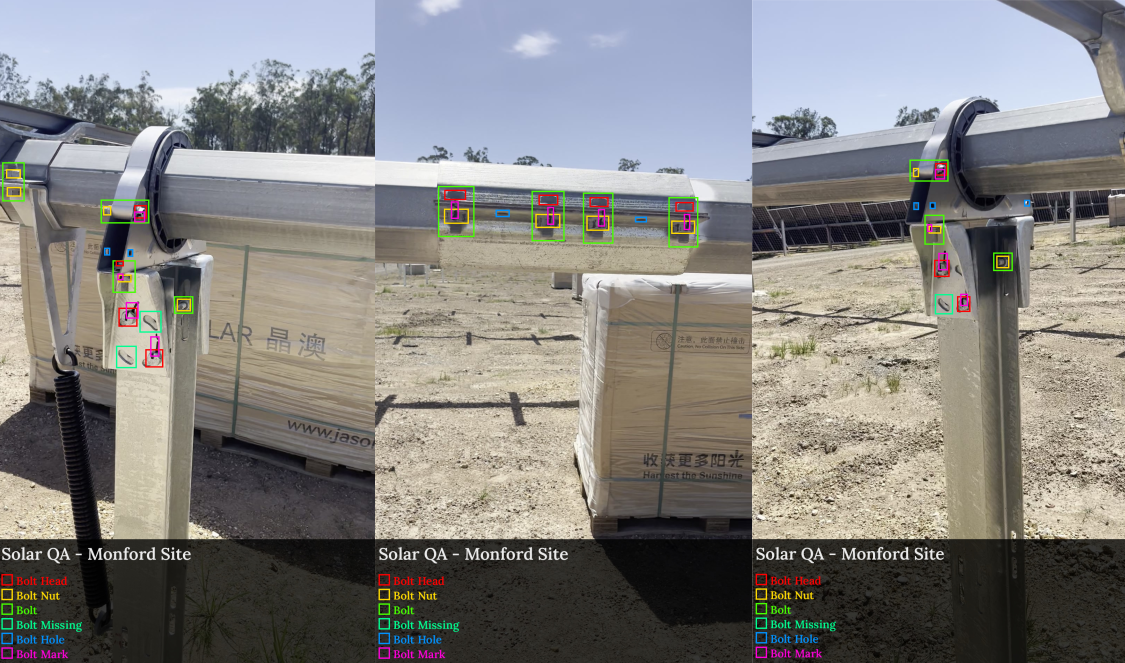}
    \caption{\centering Bolts and Nuts (Site A)}
    \label{fig:structural-and-electrical:bolts-and-nuts:1}
  \end{subfigure}
  \hfill%
  \begin{subfigure}[t]{0.32\textwidth}
    \centering
    \includegraphics[trim={21.5cm 0 0 0},clip,height=4cm]{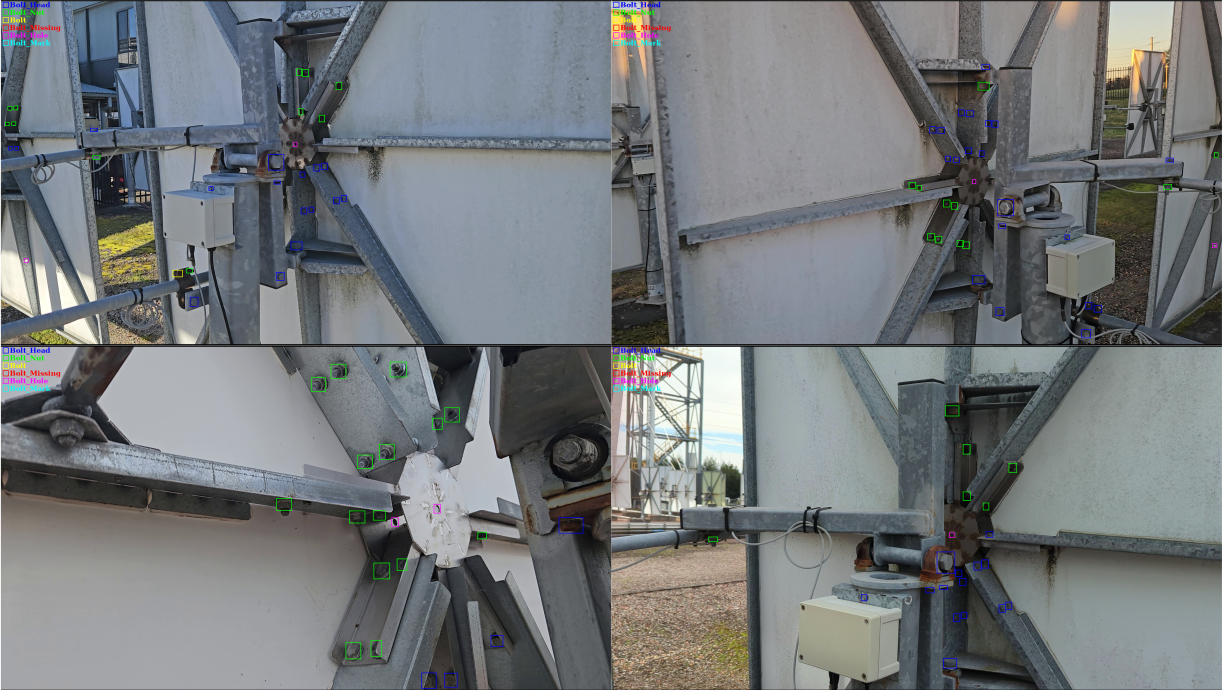}
    \caption{\centering Bolts and Nuts (Site B)}
    \label{fig:structural-and-electrical:bolts-and-nuts:1}
  \end{subfigure}
  \hfill%
  \begin{subfigure}[t]{0.32\textwidth}
    \centering
    \includegraphics[height=4cm]{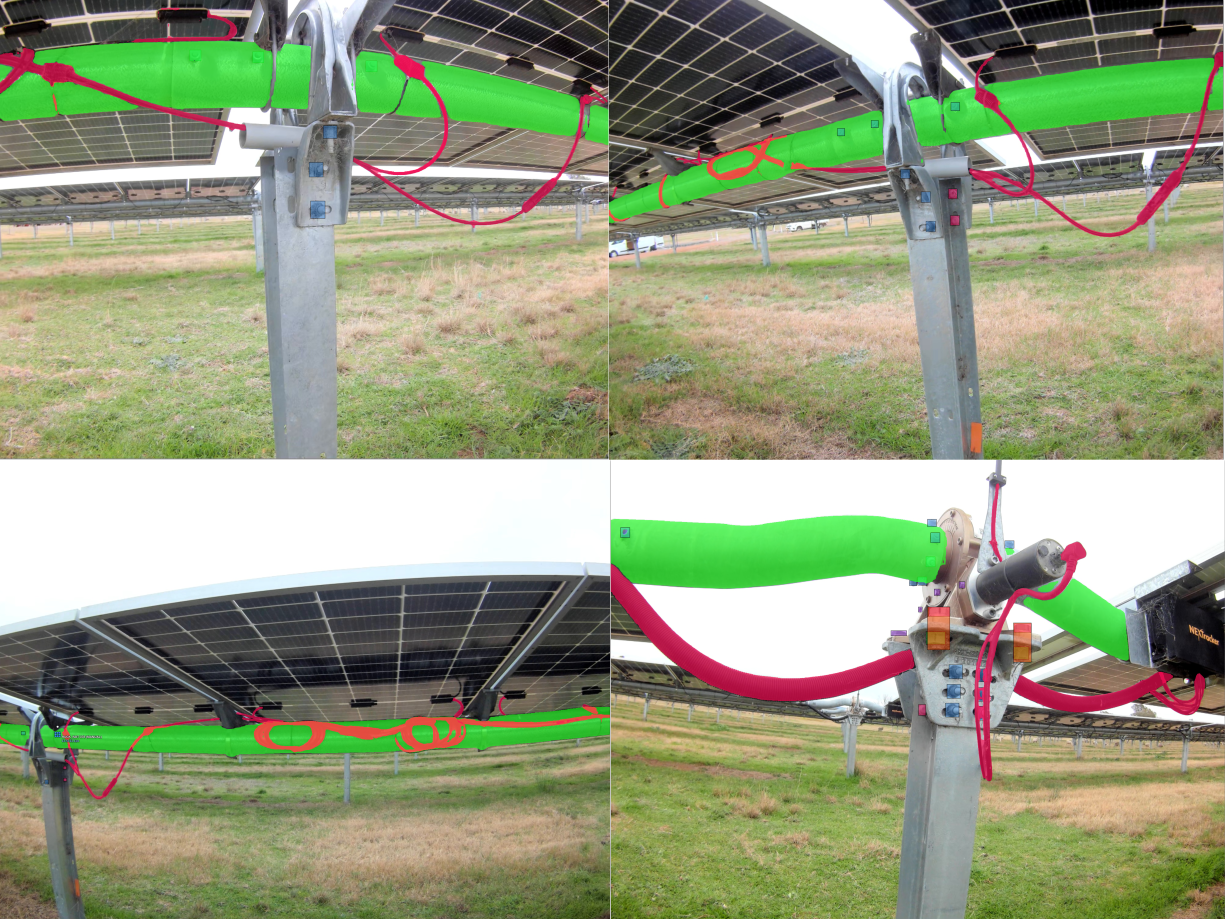}
    \caption{\centering Bolts, Nuts, Torque Tube and Wires (Site C)}
    \label{fig:structural-and-electrical:torque-tube-and-wires}
  \end{subfigure}
  \caption{Example autonomous detection of bolts, nuts, torque tube and wires.}
  \label{fig:structural-and-electrical}
\end{figure}

Structural inspection is performed by detecting bolt, nut, and hole classes within the imagery using our machine learning pipeline. The inclusion of a hole class provides additional evidence of potentially missing fasteners. Missing components can be identified by comparison with either expected structural configurations or similar structures elsewhere within the solar field. Additional ground truth permits detecting other structural and electrical components \eg{} the torque tube, distribution wires, and bolt torque marks. Example outputs are shown in Figure~\ref{fig:structural-and-electrical}.

For this work, we collected a dataset including 5K images from 3 Australian solar/industrial sites and annotated 57K bounding boxes. %
In some cases, fasteners were removed to simulate an improper installation. The Site A solar farm was freshly constructed and provided very clean samples with minimal rust/weathering. Comparatively, the Site B and Site C were significantly older with high levels of corrosion and damage.

\begin{figure}[h]
  \centering
\begin{subfigure}[t]{0.45\textwidth}
    \centering
     \includegraphics[trim={0 0 22.5cm 1.25cm},clip,width=\textwidth]{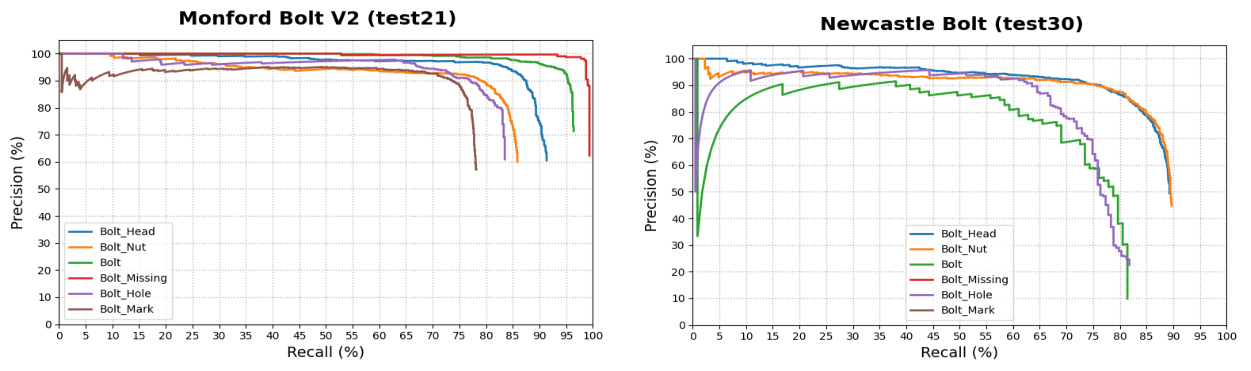}
     \caption{\centering Site A}
  \end{subfigure}
  \hfill%
  \begin{subfigure}[t]{0.45\textwidth}
    \centering
     \includegraphics[trim={22.5cm 0 0 1.25cm},clip,width=\textwidth]{figures-tricorder/Bolt-Prec-Recall.png}
    \caption{\centering Site B}
  \end{subfigure}
    \caption{Precision–recall curves for the DeNeT bolt detector.}
\label{fig:bolt-prec-recall}
\end{figure}

To quantify the accuracy of the object detector, we identified a train/test data split from the Site A/Site B datasets and plotted precision-recall curves (Figure \ref{fig:bolt-prec-recall}) for each independently. Our results indicate which classes are easiest to detect (\eg{} Bolt heads are easier to identify than torque marks) and, by comparing Site A to Site B, the effect of environmental conditions on these metrics (reduced precision and recall). 

\subsection{Electrical Inspection: Faulty MC4 connectors via thermal signature} 

MC4 connector inspection combines RGB detection, semantic segmentation, and thermal analysis. Connectors are first detected and segmented in RGB imagery using our machine learning pipeline, after which the corresponding pixels are transferred into the thermal image using the calibrated multimodal geometry. This allows connector temperature to be estimated from the segmented component region rather than from a coarse image window.

\begin{figure}[h]
  \centering
  \begin{subfigure}[t]{0.32\textwidth}
     \centering
     \includegraphics[width=\textwidth]{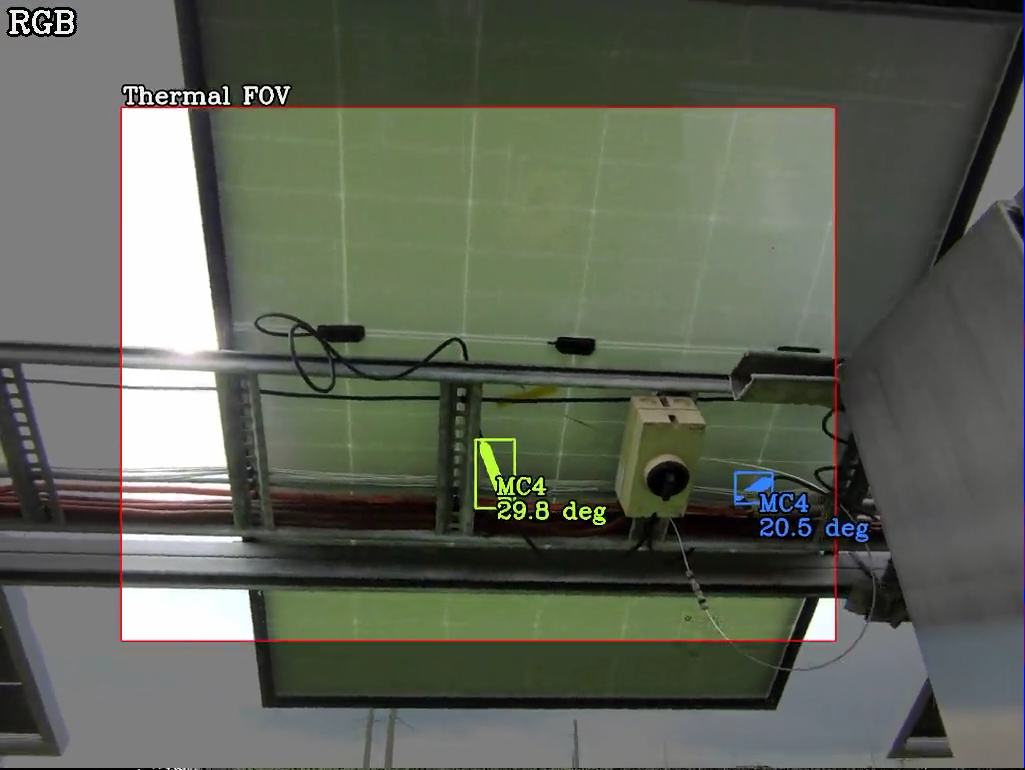}
     \caption{RGB}
     \label{fig:mc4-hot:rgb}
  \end{subfigure}%
  \hfill%
  \begin{subfigure}[t]{0.32\textwidth}
    \centering
    \includegraphics[width=\textwidth]{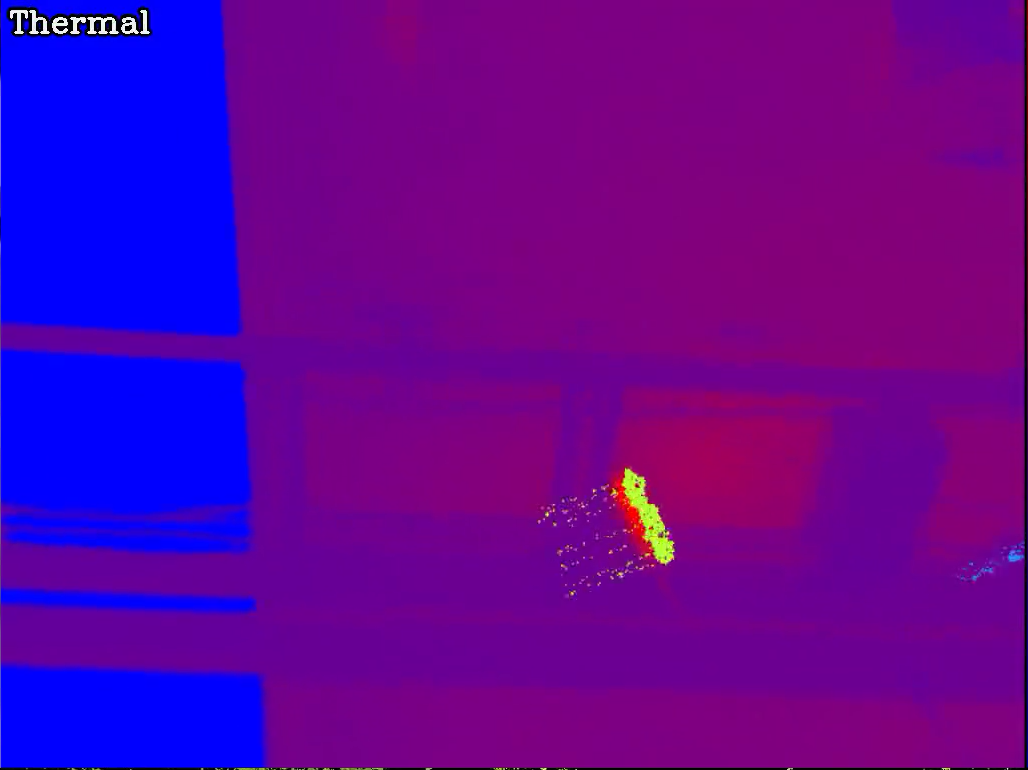}
    \caption{Thermal}
    \label{fig:mc4-hot:thermal}
  \end{subfigure}%
  \hfill%
  \begin{subfigure}[t]{0.32\textwidth}
    \centering
    \includegraphics[width=\textwidth]{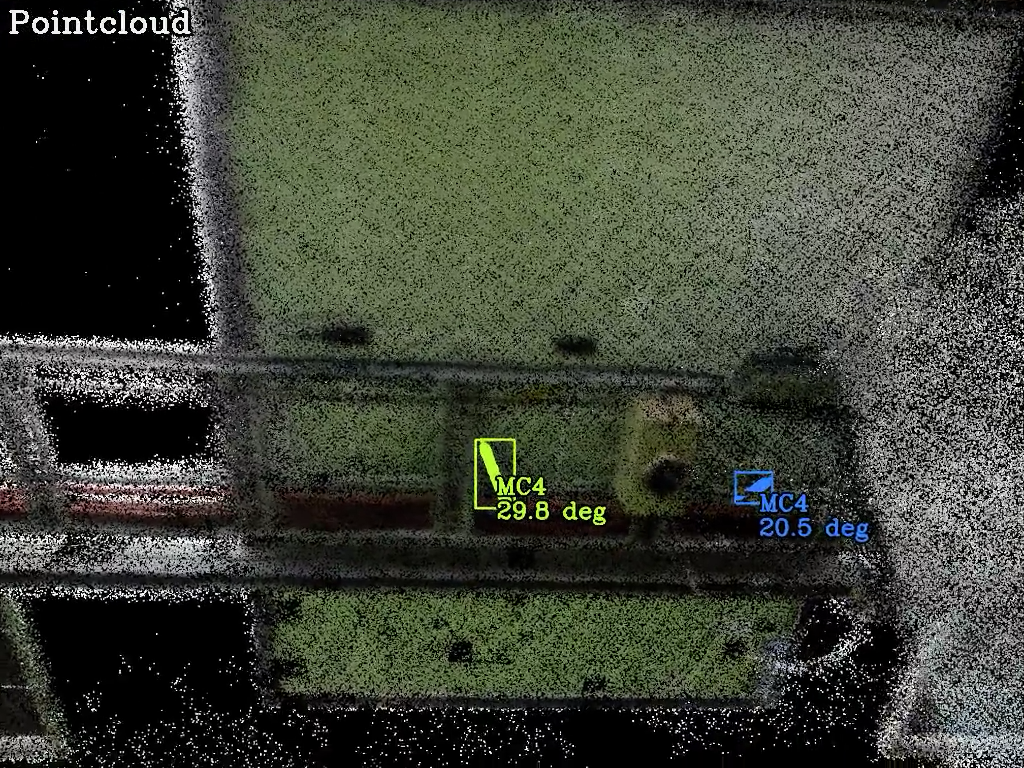}
    \caption{3D Pointcloud}
    \label{fig:mc4-hot:3d}
  \end{subfigure}%
\caption{Autonomous detection and segmentation of MC4 connectors. Note the difference in temperature between the failed connector (yellow) and the nominal connector (blue).}
\label{fig:mc4-hot}
\end{figure}

\begin{figure}[h]
  \centering
  \begin{subfigure}[t]{0.32\textwidth}
     \centering
     \includegraphics[width=\textwidth]{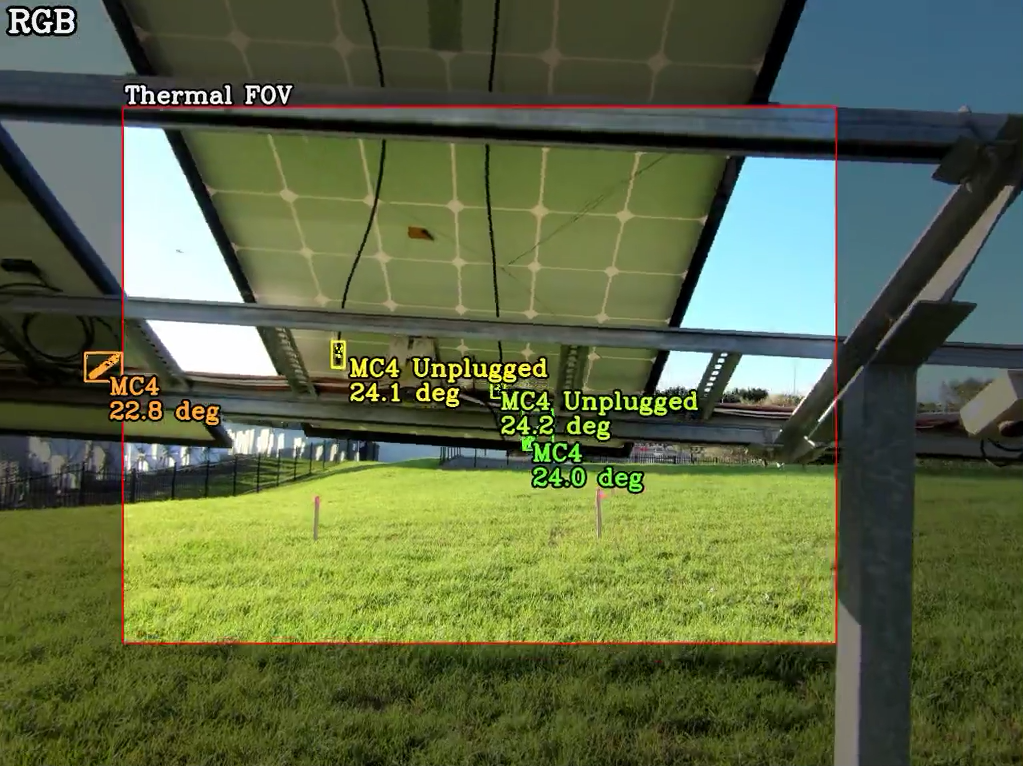}
     \caption{RGB}
     \label{fig:mc4-unplugged:rgb}
  \end{subfigure}%
  \hfill%
  \begin{subfigure}[t]{0.32\textwidth}
    \centering
    \includegraphics[width=\textwidth]{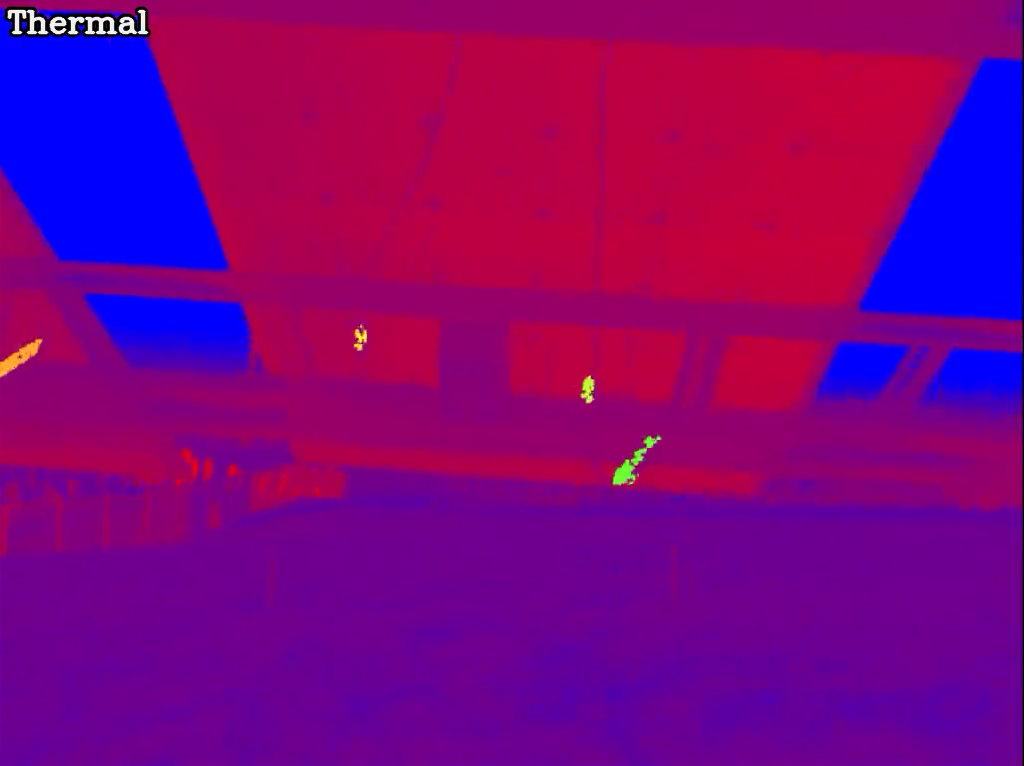}
    \caption{Thermal}
    \label{fig:mc4-unplugged:thermal}
  \end{subfigure}%
  \hfill%
  \begin{subfigure}[t]{0.32\textwidth}
    \centering
    \includegraphics[width=\textwidth]{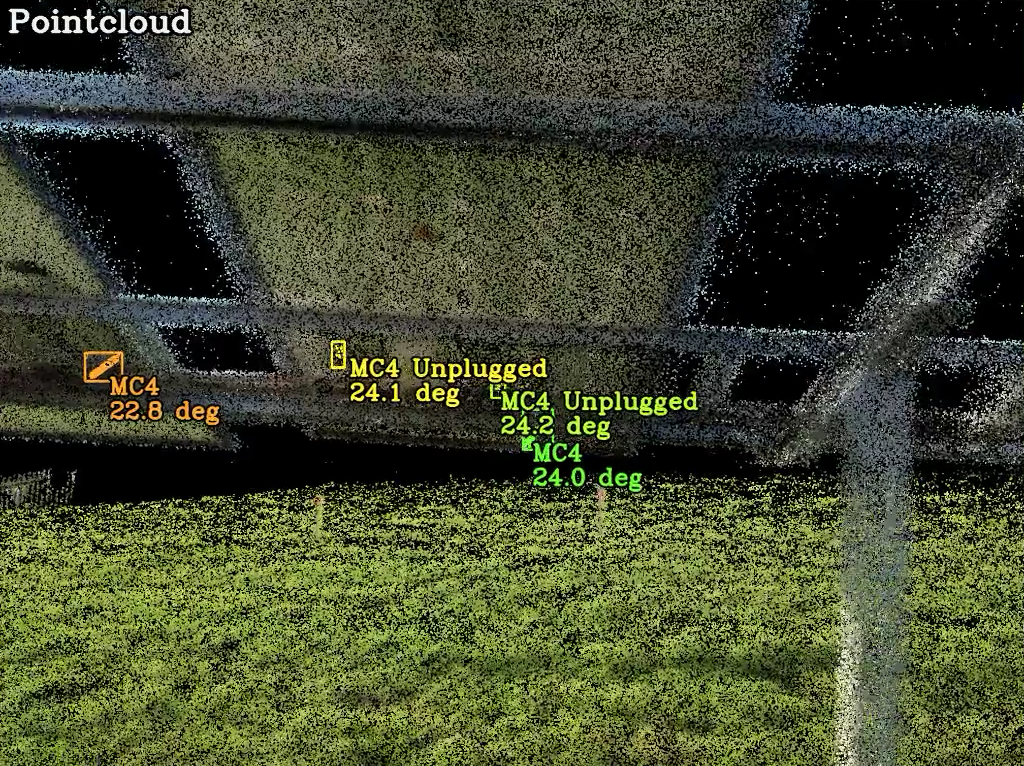}
    \caption{3D Pointcloud}
    \label{fig:mc4-unplugged:3d}
  \end{subfigure}%
\caption{Autonomous detection of connected and disconnected MC4 connectors. %
}
\label{fig:mc4-unplugged}
\end{figure}

Thermal measurements are then compared with neighbouring connectors operating under similar conditions to identify anomalous heating that may indicate increased resistance or connector degradation. Connection state is assessed separately from RGB imagery, enabling connected and disconnected MC4 connectors to be distinguished. By combining component identity, thermal response and 3D location, the inspection output can associate a candidate fault with a specific connector within the solar farm.
Examples of faulty MC4 connector detection and connected/disconnected connector-state recognition are shown in Figure~\ref{fig:mc4-hot} and Figure~\ref{fig:mc4-unplugged}.

\section{Conclusions \& Future Work}

This work demonstrates a spatially grounded approach to autonomous under-panel PV inspection, in which visual and thermal detections are associated with persistent 3D asset locations rather than treated as isolated image observations. The system combines LiDAR-inertial localisation, multimodal calibration and machine-learning-based asset recognition to identify structural and electrical components and localise candidate defects across utility-scale PV sites.

Future work will integrate the Tricorder perception stack with autonomous mobile ground robots capable of navigating beneath PV arrays and performing inspection missions without direct operator intervention. The longer-term vision is a fleet of coordinated robots that can repeatedly traverse large solar farms, build and update a shared 3D asset map, revisit previously identified components, and compare observations over time to detect emerging structural, electrical, and thermal defects. Such fleet-based inspection could complement aerial surveys with persistent under-panel coverage and enable scalable, autonomous condition monitoring across utility-scale solar farms.

\bibliographystyle{unsrt}
\bibliography{bibliography}

\end{document}